\documentclass[10pt,twocolumn,letterpaper]{article}

\usepackage{iccv}
\usepackage{times}
\usepackage{epsfig}
\usepackage{graphicx}
\usepackage{amsmath}
\usepackage{amssymb}
\usepackage[numbers,sort&compress]{natbib}
\usepackage{color}
\usepackage{multirow}
\usepackage{booktabs}
\usepackage{soul}
\usepackage[accsupp]{axessibility}

\usepackage[pagebackref=true,breaklinks=true,letterpaper=true,colorlinks,bookmarks=false]{hyperref}
\iccvfinalcopy % *** Uncomment this line for the final submission

\def\iccvPaperID{****} % *** Enter the ICCV Paper ID here

\ificcvfinal\fi

\begin{document}

%%%%%%%%% TITLE
\title{Video Adverse-Weather-Component Suppression Network \\via Weather Messenger and Adversarial Backpropagation}

% \author{Yijun Yang\\
% The Hong Kong University of Science and Technology (Guangzhou)\\
% Guangzhou, China\\
% {\tt\small 
% yyang018@connect.hkust-gz.edu.cn}
% \and

\author{
    Yijun Yang$^{1,2}$, 
    Angelica I. Aviles-Rivero$^{3}$,
    Huazhu Fu$^{4}$,  
    Ye Liu$^{5}$,
    Weiming Wang$^{6}$,
    Lei Zhu$^{1,2}$\footnotemark[2]\\
    $^1$The Hong Kong University of Science and Technology (Guangzhou)~~~\\
    $^2$The Hong Kong University of Science and Technology~~~
    $^3$University of Cambridge~~~\\
    $^4$Institute of High Performance Computing, Agency for Science, Technology and Research, Singapore~~~\\
    $^5$Tianjin University~~~
    $^6$Hong Kong Metropolitan University~~~\\
}

\maketitle

\renewcommand{\thefootnote}{\fnsymbol{footnote}} \footnotetext[2]{Lei Zhu (leizhu@ust.hk) is the corresponding author.}

%%%%%%%%% ABSTRACT
\begin{abstract}
Although convolutional neural networks (CNNs) have been proposed to remove adverse weather conditions in single images using a single set of pre-trained weights, they fail to restore weather videos due to the absence of temporal information. Furthermore, existing methods for removing adverse weather conditions (e.g., rain, fog, and snow) from videos can only handle one type of adverse weather. In this work, we propose the first framework for restoring videos from all adverse weather conditions by developing a video adverse-weather-component suppression network (ViWS-Net).
To achieve this, we first devise a weather-agnostic video transformer encoder with multiple transformer stages. Moreover, we design a long short-term temporal modeling mechanism for weather messenger to early fuse input adjacent video frames and learn weather-specific information. We further introduce a weather discriminator with gradient reversion, to maintain the weather-invariant common information and suppress the weather-specific information in pixel features, by adversarially predicting weather types. Finally, we develop a messenger-driven video transformer decoder to retrieve the residual weather-specific feature, which is spatiotemporally aggregated with hierarchical pixel features and refined to predict the clean target frame of input videos.
Experimental results, on benchmark datasets and real-world weather videos, demonstrate that our ViWS-Net outperforms current state-of-the-art methods in terms of restoring videos degraded by any weather condition.
\end{abstract}

%%%%%%%%% BODY TEXT
\section{Introduction}
 
Adverse weather conditions (including rain, fog and snow) often degrade the performance of outdoor vision systems, such as autonomous driving and traffic surveillance, by reducing environment visibility and corrupting image/video content. Removing these adverse weather effects is challenging yet a promising task. While many video dehazing/deraining/desnowing methods have been proposed, they mainly address one type of weather degradation. As they require multiple models and sets of weights for all adverse weather conditions, resulting in expensive memory and computational costs, they are unsuitable for real-time systems. Additionally, the system would have to switch between a series of weather removal algorithms, making the pipeline more complicated and less practical for real-time systems.

% Despite excellent performance in specific weather removal, these are not generic solutions for all adverse weather removal problems. This makes it difficult to adopt them for real-time systems as there have to be multiple models and sets of weights with expensive memory and computational costs. Also, the system would have to decide and switch between a series of weather removal algorithms, making the pipeline more complicated.
%
Recently, Li \etal~\cite{li2020all} proposed an All-in-One bad weather removal network that can remove any weather condition from an image, making it the first algorithm to provide a generic solution for adverse weather removal. Following this problem setting, several single-image multi-adverse-weather removal methods~\cite{valanarasu2022transweather,chen2022learning} have been developed to remove the degradation effects by one model instance of a single encoder and single decoder.
While significant progress has been witnessed for the single-image multi-adverse-weather removal task, we believe that video-level algorithms can achieve better results by utilizing the temporal redundancy from neighboring frames to reduce the inherent ill-posedness in restoration tasks. 
%\Angie{As prior information is provided, through temporal information, to reduce the inherent ill-posedness in restoration tasks.}

Therefore, a generic framework that can transform an image-level algorithm into its video-level counterpart is highly valuable. However, two bottlenecks need to be addressed: \textit{1) how to effectively maintain the temporal coherence of background details across video frames, and 2) how to prevent the perturbation of multiple kinds of weather across video frames.}

%
% To tackle the aforementioned bottlenecks, we design the first video-level algorithm to remove all adverse weather conditions with solely one set of pre-trained weights, dubbed \textbf{ViWS-Net}. Specifically, we introduce Temporally-active Weather Messenger tokens to collect weather-specific information across video frames. We also design a Long Short-term Temporal Modeling mechanism for weather messenger tokens to provide early fusion between frames and support the recovery with temporal dependences of different time spans. To impede the negative effects of multiple adverse weather conditions on the background recovery, we develop a Weather-Suppression Adversarial Learning by introducing a weather discriminator. Adversarial backpropagation is adopted between the video transformer encoder and the discriminator by gradient reversion to maintain the common background information and simultaneously suppress the weather-specific information in hierarchical pixel features. Since there has been no public dataset for video desnowing, we synthesize the first video-level snow dataset, termed KITTI-snow, which is constructed based on KITTI~\cite{liao2022kitti}. We conduct extensive experiments on video deraining, dehazing and desnowing benchmark datasets, \ie, RainMotion~\cite{wang2022rethinking}, REVIDE~\cite{zhang2021learning} and our KITTI-snow, as well as several real-world weather videos, to validate the effectiveness and generalization of our framework on video multiple adverse weather removal problem.

To tackle the aforementioned bottlenecks, we present the \textbf{Vi}deo Adverse-\textbf{W}eather-Component \textbf{S}uppression \textbf{Net}work (\textbf{ViWS-Net}), \textit{the first video-level algorithm that can remove all adverse weather conditions with only one set of pre-trained weights}. Specifically, we introduce Temporally-active Weather Messenger tokens to learn weather-specific information across video frames and retrieve them in our messenger-driven video transformer decoder. We also design a Long Short-term Temporal Modeling mechanism for weather messenger tokens to provide early fusion among frames, and support recovery with temporal dependences of different time spans. To impede the negative effects of multiple adverse weather conditions on background recovery, we develop a Weather-Suppression Adversarial Learning by introducing a weather discriminator. Adversarial backpropagation is adopted, between the video transformer encoder and the discriminator, by gradient reversion to maintain the common background information and simultaneously suppress the weather-specific information in hierarchical pixel features. 
Since there has been no public dataset for video desnowing, we synthesize the first video-level snow dataset, named KITTI-snow, which is based on KITTI~\cite{liao2022kitti}. We conduct extensive experiments on video deraining, dehazing, and desnowing benchmark datasets, including RainMotion~\cite{wang2022rethinking}, REVIDE~\cite{zhang2021learning}, and KITTI-snow, as well as several real-world weather videos, to validate the effectiveness and generalization of our framework for video multiple adverse weather removal.
Our contributions can be summarized as follows:
\begin{itemize}
    \item We propose a novel unified framework, ViWS-Net, that addresses the problem of recovering video frames from multiple types of adverse weather degradation with a single set of pre-trained weights. 
    %Our framework uses a unified architecture of a weather-agnostic video transformer encoder and a messenger-driven video transformer decoder.
    \item We introduce temporally-active weather messenger tokens that provide early temporal fusion and help retrieving the residual weather-specific information for consistent removal of weather corruptions.
    \item We design a weather-suppression adversarial learning approach that maintains weather-invariant background information and suppresses weather-specific information, thereby preventing recovery from the perturbation of various weather types.
    \item To evaluate our framework under multiple adverse weather conditions, we synthesize a video-level snow dataset KITTI-snow. Our extensive experiments on three benchmark datasets and real-world videos demonstrate the effectiveness and generalization ability of ViWS-Net. Our code is publicly available at \href{https://github.com/scott-yjyang/ViWS-Net}{https://github.com/scott-yjyang/ViWS-Net}.
\end{itemize}

\begin{figure*}[!t]
\centering
\includegraphics[width=\textwidth]{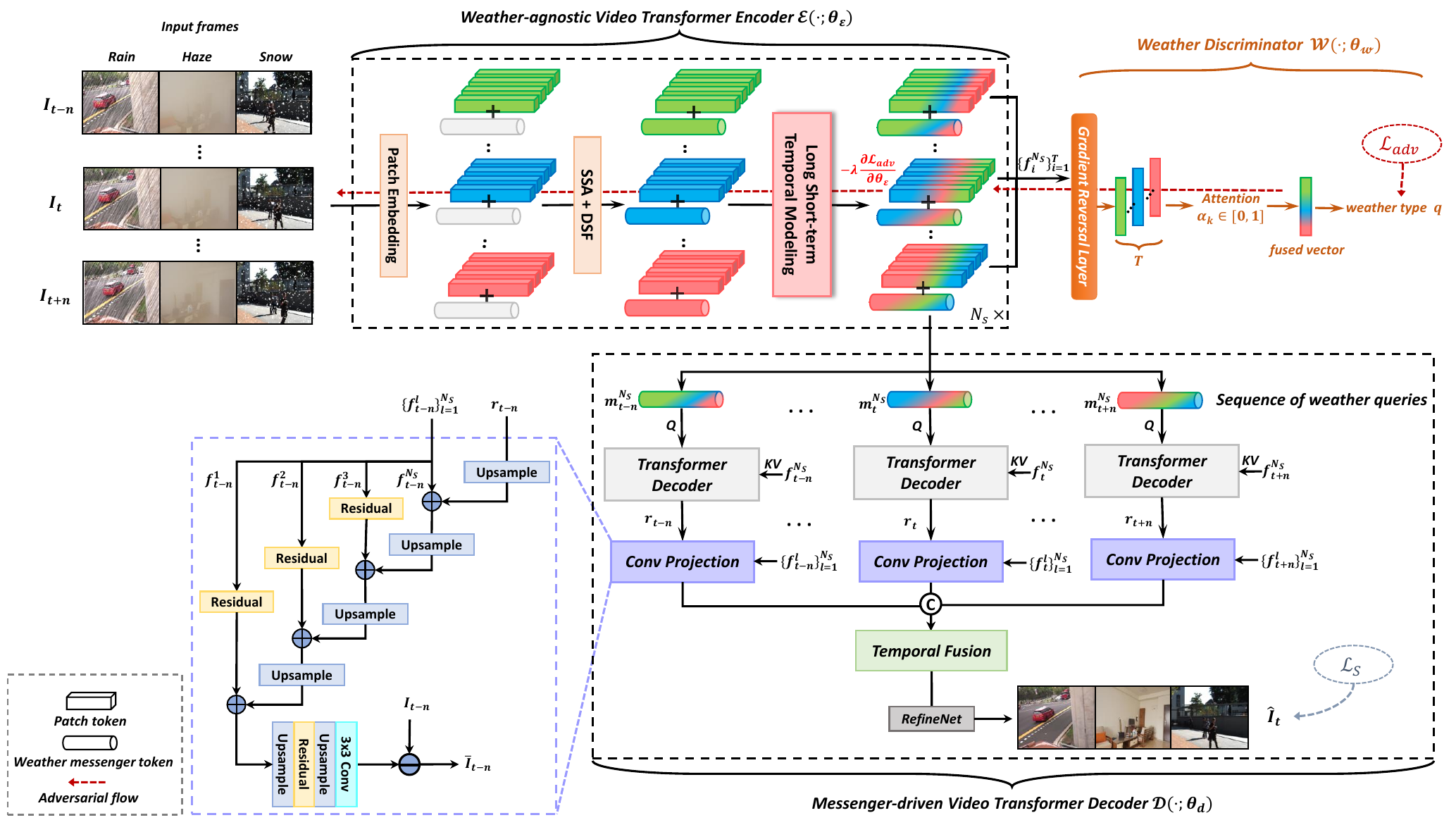}
\caption{\textbf{Overview of our ViWS-Net framework for Video Multiple Adverse Weather Removal.} Given a sequence of video frames, we divide the frames into patch tokens and concatenate them with the corresponding weather messenger token as inputs. The weather messengers temporally collect weather-specific information while the weather-agnostic video transformer encoder performs feature extraction and generates hierarchical pixel features. Simultaneously, a weather discriminator is adversarially learned by the gradient reversal layer to maintain the weather-invariant information and suppress the weather-specific counterpart. For each frame, the messenger-driven video transformer decoder leverages the last pixel feature $f^{N_s}$ as key and value, the well-learned weather messenger token $m^{N_s}$ as queries to retrieve the weather-specific feature $r$. Finally, the weather-specific feature $r$ is aggregated together with hierarchical pixel features $\{f^l\}_{l=1}^{N_s}$ across both spatial and temporal axis followed by a refinement network to obtain the final clean target frame $\hat{I}_t$.}
\label{fig:overview}
\end{figure*}

\section{Related Work}
\label{sec:relatedwork}

\noindent \textbf{Video Single-Weather Removal.}
We briefly introduce different video single-weather removal methods.
For video deraining, Garg and Nayar first modeled the video rain and develop a rain detector based on the photometric appearance of rain streak~\cite{garg2004detection,garg2006photorealistic}. Inspired by these seminal works, many subsequent methods focusing on handcrafted intrinsic priors~\cite{barnum2010analysis,bossu2011rain,brewer2008using,zhang2006rain,liu2009pixel,chen2013rain,santhaseelan2015utilizing} have been proposed in the past decades.
Recently, deep neural networks have also been employed along this research line~\cite{chen2018robust,li2018video,liu2018erase,wang2022rethinking,yan2021self,yang2019frame,yang2020self,yue2021semi}.
% Li \etal~\cite{li2018video} formulated a multi-scale convolutional sparse coding to encode the repetitive local patterns under different scales of rain streaks. 
Yang \etal~\cite{yang2019frame} built a two-stage recurrent network that utilizes dual-level regularizations toward video deraining. Wang \etal~\cite{wang2022rethinking} devised a new video rain model that accounts for rain streak motions, resulting in more accurate modeling of the rain streak layers in videos.
For video dehazing, various methods~\cite{zhang2011video,kim2012temporally,li2015simultaneous} are introduced to generate more accurate dehazed results. For example, with the development of deep learning, Ren \etal~\cite{ren2018deep} proposed a synthetic video dehazing dataset and developed a deep learning solution to accumulate information across frames for transmission estimation. To break the limit of poor performance in real-world hazy scenes, Zhang \etal~\cite{zhang2021learning} developed a video acquisition system that enabled them to capture hazy videos and their corresponding haze-free counterparts from real-world settings. Based on \cite{zhang2021learning}, Liu \etal~\cite{liu2022phase} proposed a phase-based memory network that integrates color and phase information from the current frame with that of past consecutive frames.
For snow removal, while most existing learning-based methods~\cite{chen2020jstasr,chen2021all,zhang2021deep} focused on single-image desnowing, no work explored the better solution for video desnowing using temporal information. 
%OTMS-CSC~\cite{li2021online} developed an online multi-scale convolutional sparse coding model to jointly remove rain and snow in surveillance videos. In our study, we synthesize a video desnowing dataset for further research of this line.
%
% Different from the above methods, we devise a unified single-encoder single-decoder network to tackle video all adverse weather removal problems using a single model instance.
We propose a novel approach to address the challenge of removing adverse weather effects in videos. Unlike previous methods, we adopt a unified single-encoder single-decoder network that can handle various types of adverse weather conditions using a single model instance.
%Our Weather Messenger is plugged into the encoder to gather weather information supported by temporal shifts in each video clip. Our transformer decoder is trained with these messenger tokens to learn the task-specific information to restore the clean image while the adversarial weather discriminator maintains task-invariant counterpart.

\vspace{3pt} \noindent \textbf{Single-image Multi-Adverse-Weather Removal.} 
Most recently, a body of researchers has investigated single-image multiple adverse weather removal tasks by one model instance.
Li \etal~\cite{li2020all} developed a single network-based method All-in-One with multiple task-specific encoders and a generic decoder based on Neural Architecture Search (NAS) architecture. It backpropagates the loss only to the respective feature encoder based on the degradation types. TransWeather~\cite{valanarasu2022transweather} proposed a transformer-based end-to-end network with only a single encoder and a decoder. It introduced an intra-patch transformer block into the transformer encoder for smaller weather removal. It also utilized a transformer decoder with weather type embeddings learned from scratch to adapt to different weather types. Chen \etal~\cite{chen2022learning} proposed a two-stage knowledge distillation mechanism to transfer weather-specific knowledge from multiple well-trained teachers on diverse weather types to one student model. 
%In the inference stage, the student model is used as a unified model instance to remove all the weather conditions at once.
Our study draws attention to multi-adverse-weather removal issue in videos. However, all the above methods failed to capture complementary information from temporal space. Although we can generalize them to remove adverse weather removal in a frame-by-frame manner, temporal information among video frames enables our method to work better than those image-level ones.

\vspace{3pt} \noindent \textbf{Adversarial Learning.}
% Deep learning has recently become popular due to its capacity to learn nonlinear features, which facilitates the learning of invariant features for multiple tasks.
% Adversarial learning, which is inspired by generative adversarial networks~\cite{goodfellow2020generative}, is employed in natural language processing to learn a common feature representation for multi-task learning as in~\cite{shinohara2016adversarial,liu2017adversarial,liu2018multi}.
% There are three networks in such adversarial multi-task models, \ie, a feature encoder network, a decoder network and a domain network. Based on the feature encoder network, the decoder network seeks to minimize a training loss for all the tasks, while the domain network aims to distinguish which task a data instance is from. 
Deep learning has gained popularity in recent years due to its ability to learn non-linear features, making it easier to learn invariant features for multiple tasks. Adversarial learning, inspired by generative adversarial networks~\cite{goodfellow2020generative}, has been employed in natural language processing to learn a common feature representation for multi-task learning, as demonstrated in~\cite{shinohara2016adversarial,liu2017adversarial,liu2018multi}. These adversarial multi-task models consist of three networks: a feature encoder network, a decoder network, and a domain network. The decoder network minimizes the training loss for all tasks based on the feature encoder network, while the domain network distinguishes the task to which a given data instance belongs.
Such learning paradigm has also been used to tackle the domain shift problem~\cite{ganin2015unsupervised,tzeng2017adversarial,li2018deep,shao2019multi,rahman2020correlation} to learn domain-invariant information. Inspired by those works, we further explore the common feature representation of multiple adverse weather in videos by adversarial learning paradigm.

\section{Method}
In this work, our goal is to devise the first video-level unified model to remove multiple types of adverse weather in frames with one set of model parameters. We follow an end-to-end formulation of adverse weather removal as:
\begin{equation}
    \begin{aligned} 
    \hat{I}_t &= \mathcal{D}(\mathcal{E}(\textbf{V}_i^q)),\\
    \textbf{V} = \{I_{t-n},...,&I_{t-1},I_t,I_{t+1},...,I_{t+n}\},
    \end{aligned}
\label{eq:definition}
\end{equation}
where $\textbf{V}_i^q$ is the $i$-th video clip with $T=2n+1$ frames degraded by $q$-th weather type, $\hat{I}_t$ is the recovered target frame. Different from standard image-level method All-in-One~\cite{li2020all}, our ViWS-Net tackles multiple adverse weather problem more efficiently by one video transformer encoder $\mathcal{E}(\cdot)$ and one video transformer decoder $\mathcal{D}(\cdot)$.
Next, we elaborate our solution for Video Multiple Adverse Weather Removal task.

\subsection{Overall Architecture}
The overall architecture of our ViWS-Net is displayed in Figure~\ref{fig:overview}, which consists of a weather-agnostic video transformer encoder, a messenger-driven video transformer decoder, and a weather discriminator. Without loss of generality, we build ViWS-Net based on the Shunted transformer~\cite{ren2022shunted} consisting of shunted self-attention (SSA) and detail-specific feedforward layer (DSF). 
SSA extends spatial reduction attention in PVT~\cite{wang2021pyramid} to unify multi-scale feature extractions within one self-attention layer through multi-scale token aggregation. DSF enhances local details by inserting a depth-wise convolution layer between the two fully connected layers in the feed-forward layer.

Given a sequence of video clip with $T=2n+1$ frames \{$I_{t-n},...,I_{t-1},I_t,I_{t+1},...,I_{t+n}$\} degraded by $q$-th adverse weather, our transformer encoder performs feature extraction and generates hierarchical pixel features while weather messenger tokens conduct long short-term temporal modeling for the early fusion in the temporal axis. The weather discriminator with a gradient reversal layer is adversarially learned by predicting the weather type of video clips to maintain the weather-invariant background information and suppress the weather-specific information in the pixel features. The messenger-driven video transformer decoder initializes weather type queries with temporally-active weather messenger well-learned during encoding to retrieve the residual weather-specific information from the suppressed pixel feature. Finally, the hierarchical pixel features and weather-specific feature are spatiotemporally integrated and refined to reconstruct the clean target frame. 
 %Note that $n$ is set to 2 in our setting. 
Empirically, we set $n=2$ to achieve a good trade-off between performance and computational cost.

% we first calculate the flow between the $i$-th neighbor frame and centre frame $t$ to provide more temporal information, and then warp the flow and the $i$-th neighbor frame before feeding them into the encoder. The transformer encoder performs feature extraction and generates multi-scale pyramid features. Then, the transformer decoder uses the encoded features as keys and values while using learnable weather type query embeddings as queries to disentangle weather-related information from the original feature. Our whole network is end-to-end for both training and inference.

\begin{figure}[!t]
\centering
\includegraphics[width=\columnwidth]{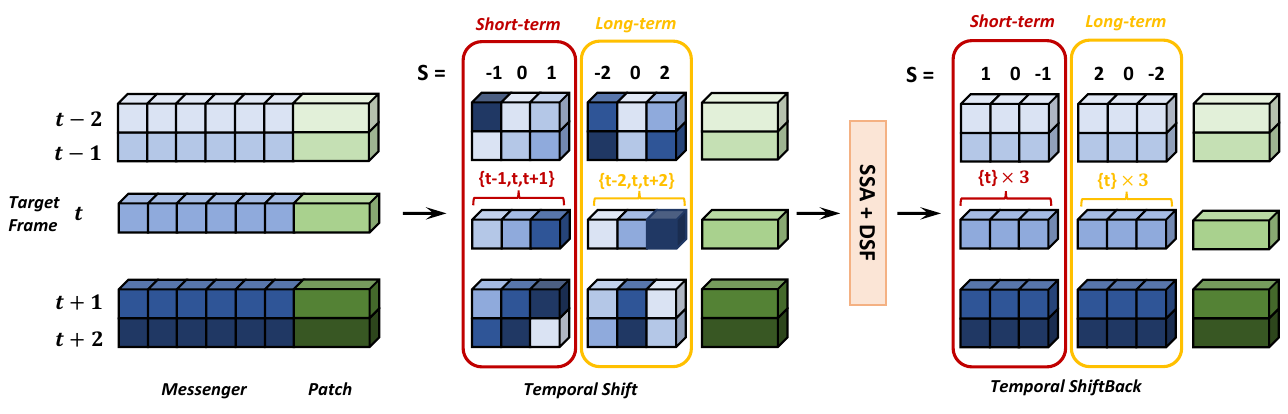}
\caption{\textbf{An illustration of our Long Short-term Temporal Modeling mechanism.} This mechanism is repeatedly applied at each stage of the transformer encoder.}
\label{fig:method_temporal}
\end{figure}

\subsection{Temporally-Active Weather Messenger}
\label{sec:mess}
Previous single-image multi-adverse-weather removal work~\cite{valanarasu2022transweather} adopted a fixed number of learnable embeddings to query weather-specific features from pixel features in the transformer decoder, termed as weather type queries. However, hindered by random initialization, they are hard to tell the robust weather-specific information during decoding. Furthermore, these query embeddings are independently learned across frames, resulting in the absence of temporal information in the video scenario. To address these limitations, we introduce weather messenger in the video transformer encoder, and the well-learned weather messengers are adopted as the weather type queries.
Specifically, a group of learnable embeddings with size of $M\times C$ is introduced as weather messenger tokens for each frame, which is denoted as $\{m^0_i\}^{T}_{i=1} \in \mathbb{R}^{T\times M\times C}$. A video clip with the resolution of $H\times W$ is divided and projected into $T\times \frac{HW}{P^2}\times C$ overlapped patch embeddings frame-by-frame, where $P$ and $C$ denote the patch size and the channel dimension respectively.
%Note that the number of embedding $M$ is set as 32. 
Then, we concatenate patch embeddings of each frame with the corresponding weather messenger tokens before feeding into the video transformer encoder:
\begin{equation}
    \{[f^0_i,m^0_i]\}^T_{i=1} \in \mathbb{R}^{T\times(\frac{HW}{P^2}+M)\times C}.
\end{equation}
The joint tokens $\{[f^0_i,m^0_i]\}^T_{i=1}$ are taken as inputs for the first stage of the transformer encoder.
Our video transformer encoder has $N_s=4$ stages and each stage consists of several blocks of SSA and DSF. The joint token of the $l$-th stage is learned as:
\begin{equation}
    \{[f^l_i,m^l_i]\}^T_{i=1}=\{DSF^l(SSA^l([f^{l-1}_i,m^{l-1}_i]))\}^T_{i=1}.
\end{equation}
Our weather messengers are temporally active between blocks of each stage to collect weather-specific information from pixel features. To further explore temporal dependence with different spans for the target frame, we conduct a long short-term temporal modeling mechanism as shown in Figure~\ref{fig:method_temporal}. 
Weather messenger tokens of one frame are separated into 6 groups and shifted along the temporal axis with different time steps (0-2) and directions (forward or backward) followed by an inverse operation (shiftback). For the target frame $I_t$, the first 3 groups model short-term dependence by shifting messenger tokens of the neighbor frames $\{I_{t-1}, I_{t+1}\}$ with one time step, while the last 3 groups model long-term dependence by shifting messenger tokens of the neighbor frames $\{I_{t-2}, I_{t+2}\}$ with two time steps. Temporal dependences of different spans endow the recovery of the target frame with the comprehensive reference of weather-specific information from past and future frames.
%With more diverse time steps and directions, Weather messenger tokens are expected to establish longer but stable temporal dependence on both past and future frames. 

\subsection{Weather-Suppression Adversarial Learning}
To construct a weather-agnostic transformer encoder, inspired by domain adaptation~\cite{ganin2015unsupervised}, we design Weather-Suppression Adversarial Learning to learn a great feature space maintaining weather-invariant background information and suppressing weather-specific information. To this end, we optimize a weather discriminator for classifying the weather types by adversarial backpropagation.
%
%\ie, to focus on learning features that combine discriminativeness and weather-invariance. 
%This objective is achieved by jointly optimizing the query-driven transformer decoder and weather predictor. 

%
%The parameters of the transformer encoder are simultaneously optimized in order to minimize the loss of the query-driven transformer decoder and to maximize the loss of weather discriminator.
%
Notably, a gradient reversal layer (GRL) is inserted between the video transformer encoder and weather discriminator. During backpropagation, GRL takes the gradient from the weather discriminator, multiplies it by $-\lambda$ and passes it to the transformer encoder. 
To predict the weather type of a video clip, we combine information from all frames of one video clip by computing an attention-weighted average of their vector representations. We apply the gated attention mechanism by using the sigmoid function to provide a learnable non-linearity that increases model flexibility. An attention score $\alpha_i$ is computed on each frame as:
\begin{equation}
    \alpha_i = \frac{\exp{\{\textbf{w}_1^T(tanh(\textbf{w}_2\textbf{v}_i^T)\cdot sigm(\textbf{w}_3\textbf{v}_i^T)}\}}
                {\sum_{k=1}^T \exp{\{\textbf{w}_1^T(tanh(\textbf{w}_2\textbf{v}_k^T)\cdot sigm(\textbf{w}_3\textbf{v}_k^T)}\}},
\end{equation}
where $\textbf{w}_1$, $\textbf{w}_2$, $\textbf{w}_3$ are learnable parameters.
This process yields an attention-weighted fused vector representation, which reads:
\begin{equation}
    \textbf{v} = \sum_{i=1}^{T} \alpha_i \textbf{v}_i,
\end{equation}
where $\textbf{v}_i$ is the vector representation from feature embeddings of the frame $i$. The weather type is finally obtained from the fused vector by one fully connected layer.
While the weather discriminator $\mathcal{W}(\cdot)$ seeks an accurate prediction for weather types, the video transformer encoder strives to generate weather-agnostic pixel features.
%misleading the discriminator.
%
The adversarial loss can be thus achieved by min-max optimization as:
\begin{equation}
    \mathcal{L}_{adv} = \min_{\theta_w}\left(\lambda \max_{\theta_\varepsilon}(\sum_{q=1}^Q \sum_{i=1}^{N_q}q \log[\mathcal{W}(\mathcal{E}(\textbf{V}_i^q)])\right).
\end{equation}
%where $\textbf{V}_i^p$ is the $i$-th video clip degraded by $p$-th weather type.

%Our weather-suppression adversarial learning develops from the basic idea that the weather discriminator $\mathcal{D}$ tries its best to make a correct classification on the weather type while the transformer encoder $\mathcal{E}$ impedes its advance. 
%At the training stage, the adversarial loss would finally reach a compromised point where the transformer encoder and weather discriminator make a tie. By simultaneously optimizing the messenger-driven video transformer decoder as Section~\ref{sec:mess} introduces, both weather-specific and weather-invariant features can be learned for better weather removal. 
Our weather-suppression adversarial learning develops from the basic idea that the weather-specific information is suppressed in hierarchical pixel features in the transformer encoder by downplaying the discrimination of weather types. This protects the recovery of the target frame from perturbations by different weather types, and thus concentrates the model on the weather-invariant background information. At the training stage, weather-suppression adversarial learning is applied to empower the video transformer encoder with the characteristic of weather-agnostic. At the inference stage, video frames are only fed into the video transformer encoder and decoder for weather removal.

\subsection{Messenger-driven Video Transformer Decoder}

Intuitively, while weather-suppression adversarial learning largely impedes the appearance of weather-specific information, the residual still may exist in pixel features when the adversarial loss reaches a saddle point.
To localize the perturbation from the residual weather-specific information, we design Messenger-driven Video Transformer Decoder to retrieve such information and recover frames from hierarchical features using temporally-active weather messengers described in Section~\ref{sec:mess}. 
%
%Firstly, The encoded feature of each frame is respectively fed into the transformer decoder to extract weather-specific information. Instead of randomly initializing and independently learning query embeddings in \cite{valanarasu2022transweather}, 
Firstly, we adopt the well-learned weather messengers $\{m_i^{N_S}\}_{i=1}^T$ to query the residual weather-specific information. After long short-term temporal modeling in the transformer encoder, weather messengers are trained to locate more true positives of adverse weather in pixel features referring to rich temporal information, than independently-learned query embeddings in \cite{valanarasu2022transweather}.
%
%Unlike $Q, K, V$ are projected from the same input in traditional self-attention block~\cite{vaswani2017attention}, $Q$ here is the copycat of our well-learned weather messenger while $K, V$ are taken from the last feature of the transformer encoder $\{f_i^{N_S}\}_{i=1}^T$. 
With the pixel feature $\{f_i^{N_S}\}_{i=1}^T$ as key and value, the transformer decoder generates the weather-specific feature $\{r_i\}_{i=1}^T$. Note that the transformer decoder here operates at a single stage but has multiple blocks, which are similar to the stage of the transformer encoder. 
As illustrated in Figure~\ref{fig:overview}, the weather-specific feature is spatially integrated with hierarchical pixel features in the convolution projection block with pairs of an upsampling layer and a 2D convolution residual layers frame-by-frame. To recover details of the background, we subtract the outputs from the original frames.
%Hierarchical features from the transformer encoder and weather-specific feature from the transformer decoder are integrated and refined by our Spatiotemporal Refinement Block (SRB). Our SRB block starts from four 2D convolutional layers to achieve spatial integration. 
After that, we concatenate the outputs of frames and feed them into the temporal fusion block consisting of three consecutive 3D convolution layers to achieve temporal integration. Finally, we obtain the clean target frame $\hat{I}_t$ by applying a refinement network, which is a vanilla and much smaller version of our ViWS-Net, onto the initial recovered results with tiny artifacts.

The supervised objective function is composed of a smooth L1 loss and a perceptual loss as follows: 
\begin{align}
 \mathcal{L}_{S} &=  \mathcal{L}_{smoothL_1} + \gamma_1\mathcal{L}_{perceptual},  
\label{eq:suploss}  \; \text{with} \\
\mathcal{L}_{smoothL_1}& = \begin{cases}
	      0.5(\hat{I}_t-B_t)^2, & if |\hat{I}_t-B_t|<1 \\
	      |\hat{I}_t-B_t|-0.5, & otherwise,
		   \end{cases}
\label{eq:smooth}\\
     \mathcal{L}_{perceptual}\!& = \!\mathcal{L}_{mse}(VGG_{3,8,15}(\hat{I}_t),VGG_{3,8,15}(B_t)),\!
\label{eq:percpetual} 
\end{align}
where $\hat{I}_t,B_t$ denote the prediction and ground truth of the target frame, respectively.
The overall objective function is composed of supervised loss and adversarial loss, which can be defined as follows:
\begin{equation}
\mathcal{L}_{total} = \mathcal{L}_{S} + \gamma_2\mathcal{L}_{adv},
\label{eq:loss}
\end{equation}
where $\gamma_1$ and $\gamma_2$ are the balancing hyper-parameters, empirically set as 0.04 and 0.001, respectively.

%------------------------------------------------------------------------
%%%% Experiments

\section{Experiments}
In this section, we describe in detail the range of experiments that we conducted to validate our proposed method.

\begin{table}[!t]
\centering
  \caption{The data statistics of RainMotion, REVIDE and KITTI-snow for our video multiple adverse weather removal. The mixed training set is composed of the training set from the three datasets.}
    \resizebox{1.0\columnwidth}{!}{%
\begin{tabular}{c|c|c|c|c|c} 
\toprule
\textbf{Weather}      & \textbf{Dataset}            & \textbf{Split} & \textbf{Video Num} & \textbf{Video Length} & \textbf{Video Frame Num}  \\ 
\hline
\multirow{2}{*}{Rain} & \multirow{2}{*}{RainMotion} & train          & 40                 & 50                    & 2000                      \\ 
\cline{3-6}
                      &                             & test           & 40                 & 20                    & 800                       \\ 
\hline
\multirow{2}{*}{Haze} & \multirow{2}{*}{REVIDE}     & train          & 42                 & 7-34                  & 928                       \\ 
\cline{3-6}
                      &                             & test           & 6                  & 20-31                 & 154                       \\ 
\hline
\multirow{2}{*}{Snow} & \multirow{2}{*}{KITTI-snow} & train          & 35                 & 50                    & 1750                      \\ 
\cline{3-6}
                      &                             & test           & 15                 & 50                    & 750                       \\
\bottomrule
\end{tabular}
    }
\label{tab:dataset}
\end{table}

\begin{table*}[!tbp]
\centering
  \caption{\textbf{Quantitative evaluation for video multiple adverse weather removal.} For Original Weather, these methods are trained on the weather-specific training set and tested on the weather-specific testing set. For Rain, Haze, and Snow, these methods are trained on a mixed training set and tested on the weather-specific testing set. The average performance is calculated on Rain, Haze, and Snow. PSNR and SSIM are adopted as our evaluation metrics. The top values are denoted in red.}
    \resizebox{1.0\textwidth}{!}{%
\begin{tabular}{c|c|l|c|cc|cccccccc} 
\toprule
\multicolumn{2}{c|}{\multirow{2}{*}{\textbf{Methods }}}    & \multicolumn{1}{c|}{\multirow{2}{*}{\textbf{Type}}} & \multirow{2}{*}{\textbf{Source }}            & \multicolumn{10}{c}{\textbf{Datasets }}                                                                                                                                                                                                                                                                                                 \\ 
\cline{5-14}
\multicolumn{2}{c|}{}                                      & \multicolumn{1}{c|}{}                               &                                              & \multicolumn{2}{c|}{\textbf{Original Weather }} & \multicolumn{2}{c}{\textbf{Rain }}                                  & \multicolumn{2}{c}{\textbf{Haze }}                                 & \multicolumn{2}{c}{\textbf{Snow }}                                 & \multicolumn{2}{c}{\textbf{Average}}                                  \\ 
\hline
\hline
\multirow{4}{*}{\textbf{Derain}}      & \textbf{PReNet}~\cite{ren2019progressive}    & Image                                      & CVPR'19                                      & 27.06 & 0.9077                                  & 26.80                           & 0.8814                            & 17.64                           & 0.8030                           & 28.57                           & 0.9401                           & 24.34~                           & 0.8748~                            \\
                                      & \textbf{SLDNet}~\cite{yang2020self}    & Video                                      & CVPR'20                                      & 20.31 & 0.6272                                  & 21.24                           & 0.7129                            & 16.21                           & 0.7561                           & 22.01                           & 0.8550                           & 19.82~                           & 0.7747~                            \\
                                      & \textbf{S2VD}~\cite{yue2021semi}      & Video                                      & CVPR'21                                      & 24.09 & 0.7944                                  & 28.39                           & 0.9006                            & 19.65                           & 0.8607                           & 26.23                           & 0.9190                           & 24.76~                           & 0.8934~                            \\
                                      & \textbf{RDD-Net}~\cite{wang2022rethinking}   & Video                                      & ECCV'22                                      & 31.82 & 0.9423                                  & 30.34                           & 0.9300~                           & 18.36                           & 0.8432                           & 30.40~                          & 0.9560~                          & 26.37~                           & 0.9097~                            \\ 
\hline
\multirow{4}{*}{\textbf{Dehaze }}     & \textbf{GDN}~\cite{liu2019griddehazenet}       & Image                                      & ICCV'19                                      & 19.69 & 0.8545                                  & 29.96                           & 0.9370~                           & 19.01                           & 0.8805                           & 31.02                           & 0.9518                           & 26.66~                           & 0.9231~                            \\
                                      & \textbf{MSBDN}~\cite{dong2020multi}     & Image                                      & CVPR'20                                      & 22.01 & 0.8759                                  & 26.70                           & 0.9146~                           & 22.24                           & 0.9047                           & 27.07                           & 0.9340                           & 25.34~                           & 0.9178~                            \\
                                      & \textbf{VDHNet}~\cite{ren2018deep}    & Video                                      & TIP'19                                       & 16.64 & 0.8133                                  & 29.87                           & 0.9272~                           & 16.85                           & 0.8214                           & 29.53                           & 0.9395                           & 25.42~                           & 0.8960~                            \\
                                      & \textbf{PM-Net}~\cite{liu2022phase}    & Video                                      & MM'22                                        & 23.83 & 0.8950~                                 & 25.79                           & 0.8880~                           & 23.57                           & 0.9143                           & 18.71                           & 0.7881                           & 22.69~                           & 0.8635~                            \\ 
\hline
\multirow{4}{*}{\textbf{Desnow }}     & \textbf{DesnowNet}~\cite{liu2018desnownet} & Image                                     & TIP'18                                       & 28.30 & 0.9530                                  & 25.19                           & 0.8786                            & 16.43                           & 0.7902                           & 27.56                           & 0.9181                           & 23.06~                           & 0.8623~                            \\
                                      & \textbf{DDMSNET}~\cite{zhang2021deep}   & Image                                      & TIP'21                                       & 32.55 & 0.9613                                  & 29.01                           & 0.9188~                           & 19.50                           & 0.8615                           & \textbf{\textcolor{red}{32.43}}                           & \textbf{\textcolor{red}{0.9694}}                           & 26.98~                           & 0.9166~                            \\
                                      & \textbf{HDCW-Net}~\cite{chen2021all}  & Image                                      & ICCV'21                                      & 31.77 & 0.9542                                  & 28.10~                          & 0.9055~                           & 17.36                           & 0.7921                           & 31.05                           & 0.9482~                          & 25.50~                           & 0.8819~                            \\
                                      & \textbf{SMGARN}~\cite{cheng2022snow}    & Image                                      & \textcolor[rgb]{0.141,0.161,0.184}{TCSVT‘22} & 33.24 & 0.9721                                  & 27.78                           & 0.9100~                           & 17.85                           & 0.8075                           & 32.34                           & 0.9668                           & 25.99~                           & 0.8948~                            \\ 
\hline
\multirow{4}{*}{\textbf{Restoration}} & \textbf{MPRNet}~\cite{zamir2021multi}    & Image                                      & CVPR’21                                      & — —    & — —                                      & 28.22~                          & 0.9165~                           & 20.25                           & 0.8934                           & 30.95                           & 0.9482~                          & 26.47~                           & 0.9194~                            \\
                                      & \textbf{EDVR}~\cite{wang2019edvr}      & Video                                      & CVPR'19                                      & — —    & — —                                      & 31.10~                          & 0.9371~                           & 19.67                           & 0.8724                           & 30.27                           & 0.9440~                          & 27.01~                           & 0.9178~                            \\
                                      & \textbf{RVRT}~\cite{liang2022recurrent}      & Video                                      & NIPS'22                                      & — —    & — —                                      & 30.11~                          & 0.9132~                           & 21.16                           & 0.8949                           & 26.78                           & 0.8834                           & 26.02~                           & 0.8972~                            \\
                                      & \textbf{RTA}~\cite{zhou2022revisiting}       & Video                                      & CVPR'22                                      & — —    & — —                                      & 30.12~                          & 0.9186~                           & 20.75                           & 0.8915                           & 29.79                           & 0.9367                           & 26.89~                           & 0.9156~                            \\ 
\hline
% All-in-one  & 26.62 & 0.8948 & 20.88 & 0.9010 & 30.09 & 0.9431 & 25.86 & 0.9130  \\
\multicolumn{2}{c|}{\textbf{All-in-one}~\cite{li2020all}}                & Image                                      & CVPR‘20                                      & — —    & — —                                      & 26.62                           & 0.8948~                           & 20.88                           & 0.9010                           & 30.09                          & 0.9431                           & 25.86~                           & 0.9130~                            \\
\multicolumn{2}{c|}{\textbf{UVRNet}~\cite{kulkarni2022unified}}                         & Image                                      & TMM‘22                                      & — —    & — —                                      & 22.31                           & 0.7678~                           & 20.82                           & 0.8575                           & 24.71                           & 0.8873                           & 22.61~                           & 0.8375~                            \\
\multicolumn{2}{c|}{\textbf{TransWeather}~\cite{valanarasu2022transweather}}                & Image                                      & CVPR‘22                                      & — —    & — —                                      & 26.82                           & 0.9118~                           & 22.17                           & 0.9025                           & 28.87                           & 0.9313                           & 25.95~                           & 0.9152~                            \\
\multicolumn{2}{c|}{\textbf{TKL}~\cite{chen2022learning}}                         & Image                                      & CVPR‘22                                      & — —    & — —                                      & 26.73                           & 0.8935~                           & 22.08                           & 0.9044                           & 31.35                           & 0.9515                           & 26.72~                           & 0.9165~                            \\
% UVRNet  & 22.31	& 0.7678 & 20.82 & 0.8575 & 24.71 & 0.8873 & 22.61 & 0.8375  \\
\multicolumn{2}{c|}{\textbf{Ours }}                        & Video                                      & — —                                           & — —    & — —                                      & \textbf{\textcolor{red}{31.52}} & \textbf{\textcolor{red}{0.9433~}} & \textbf{\textcolor{red}{24.51}} & \textbf{\textcolor{red}{0.9187}} & 31.49 & 0.9562 & \textbf{\textcolor{red}{29.17~}} & \textbf{\textcolor{red}{0.9394~}}  \\
%\multicolumn{2}{c|}{\textbf{Ours }}                        & Video                                      & — —                                           & — —    & — —                                      & \textbf{\textcolor{red}{31.42}} & \textbf{\textcolor{red}{0.9437~}} & \textbf{\textcolor{red}{24.53}} & \textbf{\textcolor{red}{0.9198}} & \textbf{\textcolor{red}{31.38}} & \textbf{\textcolor{red}{0.9550}} & \textbf{\textcolor{red}{29.11~}} & \textbf{\textcolor{red}{0.9395~}}  \\

\bottomrule
\end{tabular}
    }
\label{tab:main}
\end{table*}

\subsection{Datasets}
Various video adverse weather datasets are used in our experiments. Table~\ref{tab:dataset} summarizes the information of our video multiple adverse weather datasets. RainMotion~\cite{wang2022rethinking} is the latest video deraining dataset synthesized based on NTURain~\cite{chen2018robust}. It has five large rain streak masks, making it more demanding to remove the rain streaks. REVIDE~\cite{zhang2021learning} is the first real-world video dehazing dataset with high fidelity real hazy conditions recording indoor scenes.
To our best knowledge, there have not been any public video-level snow datasets yet. Thus, we built our own video desnowing dataset named KITTI-snow. The details of KITTI-snow are presented as follows. At the training stage, we merge the training set of the three datasets to learn a unified model. For the testing stage, we evaluate our model on three testing sets, respectively.

\begin{figure}[!t]
\centering
\includegraphics[width=\columnwidth]{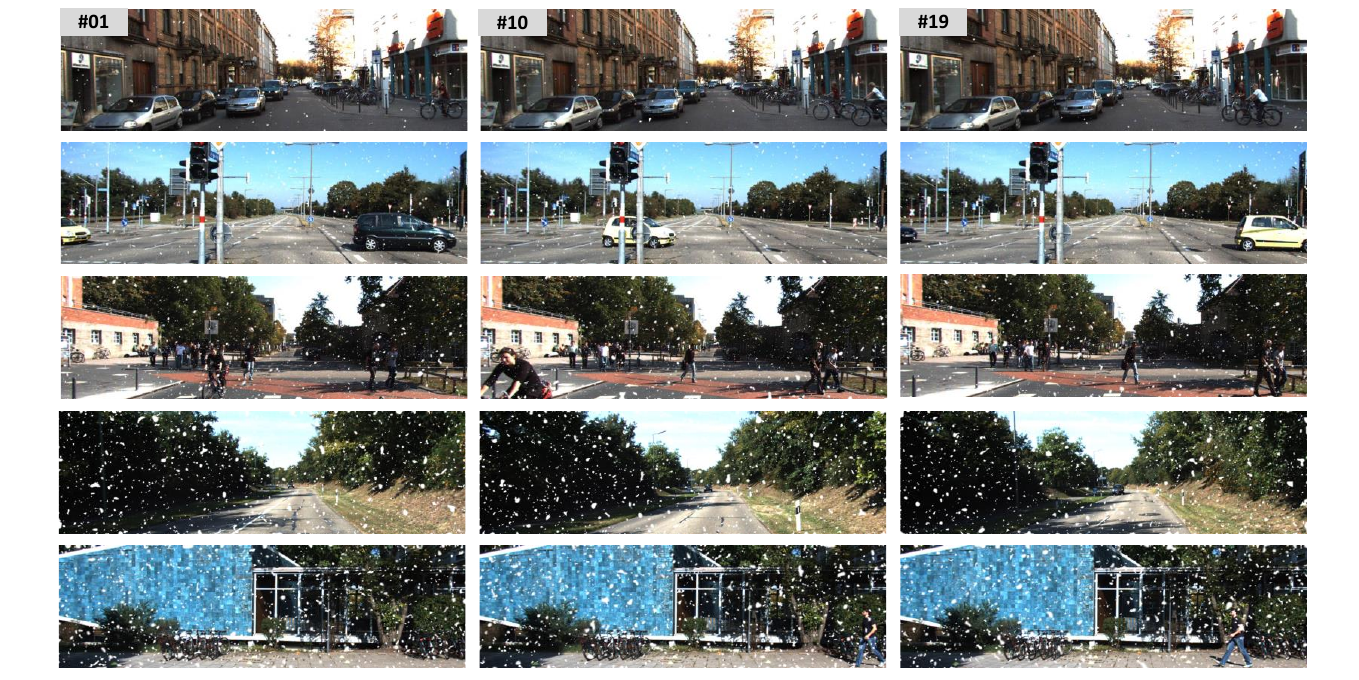}
\caption{\textbf{Example frames of five synthesized videos in KITTI-snow.} The snowflakes in each video are sampled from different distributions.}
\label{fig:kitti_snow}
\vspace{-1mm}
\end{figure}

\noindent \textbf{KITTI-snow:}
We create a synthesized outdoor dataset called KITTI-snow that comprises 50 videos with a total of 2500 frames, all featuring snowy conditions. 
%
% Specifically, we randomly collected two groups of videos from KITTI~\cite{liao2022kitti}, while the first group contains 35 videos as the training set and the second group contains 15 videos as the testing set.
Specifically, we randomly collect two groups of videos from KITTI~\cite{liao2022kitti}. The first group consists of 35 videos and is treated as the training set, while the second group includes 15 videos and is treated as the testing set.
Given each clean video, we synthesize snowflakes with different properties (i.e. transparency, size and position) according to Photoshop's snow synthesis tutorial. To better simulate the real-world snow scene, gaussian blurring is applied onto snow particles. 
To model the temporal consistency, we sample the position, size and blurring degree of snow in different frames of the same video from the same distribution. 
%The random seeds of each frame are different, making the diversity of video frames.
%It is worth noting that the position, size and blurring degree of snow in different frames of the same scene are different.
The spatial resolution of video frames is $1000 \times 300$. Figure~\ref{fig:kitti_snow} presents the example frames of five videos with different distributions in our synthetic dataset.

\begin{figure*}
\centering
\includegraphics[width=\textwidth]{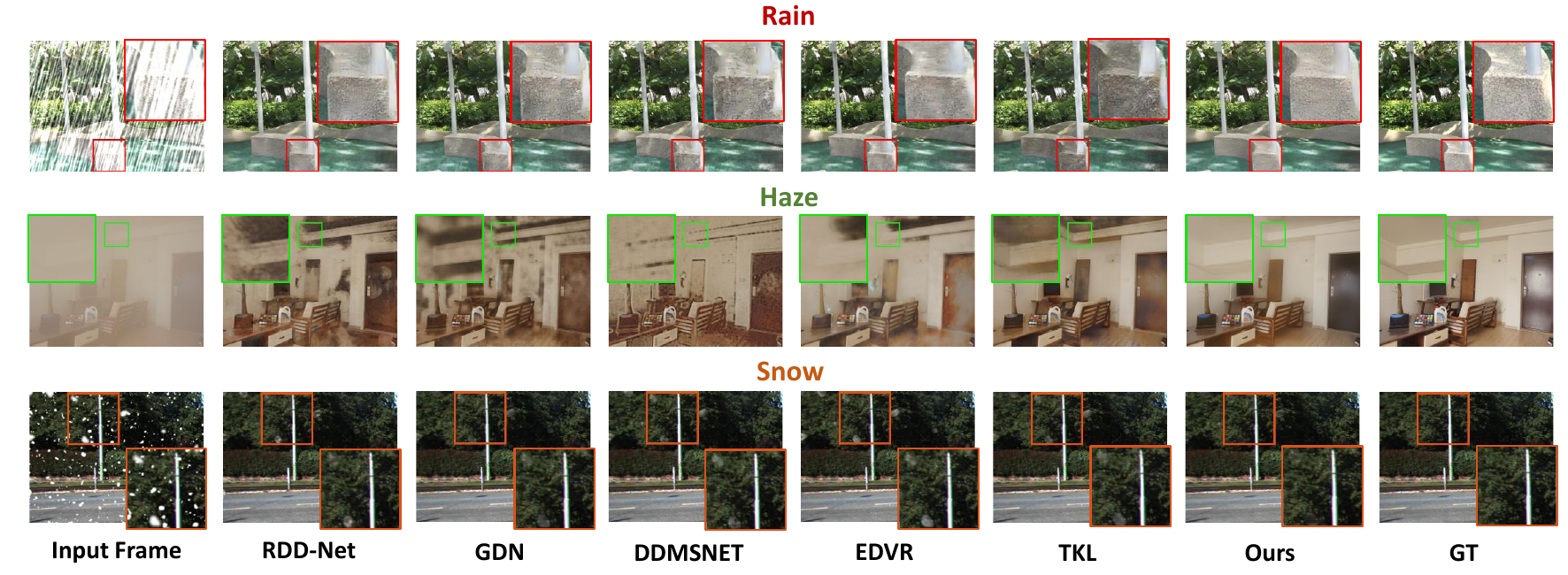}
\caption{\textbf{Qualitative Comparison between adverse weather removal algorithms.} The best algorithms designed for different tasks are selected to present the results on the example frames degraded by rain, haze, snow, respectively. The color box indicates the detailed comparison of weather removal.}
\label{fig:vis_comparison}
\end{figure*}

\subsection{Implementation Details}
For training details, the proposed framework was trained on two NVIDIA RTX 3090 GPUs and implemented on the Pytorch platform. Our framework is empirically trained for 500 epochs in an end-to-end way and the Adam optimizer is applied. The initial learning rate is set to $2 \times 10^{-4}$ and decayed by 50\% every 100 epochs. We randomly crop the video frames to $224 \times 224$.
We empirically set $n=2$, which means that our network receives 5 frames for each video clip. A batch of 12 video clips evenly composed of three weather types (\ie, rain, haze, snow) is fed into the network for each time. 
%The average running time of our network is about 0.1130s for one video frame with a resolution of 640×480.

For method details, the number of weather messenger tokens  $M$ for each frame is set to 48.
%
%The balancing hyper-parameters for loss function, $\gamma_1, \gamma_2$, are empirically set to 0.04, 0.001, respectively.
%
In order to suppress noisy signal from the weather discriminator at the early stages of the training procedure, we gradually change the adaptation factor $\lambda$ from 0 to 1 following the schedule:
\begin{equation}
        \lambda = \frac{2}{1+\exp(-10\cdot p)}-1,
\label{eq:lambda}
\end{equation}
where $p$ is the current iteration number divided by the total iteration number.

\begin{figure}
\centering
\includegraphics[width=\columnwidth]{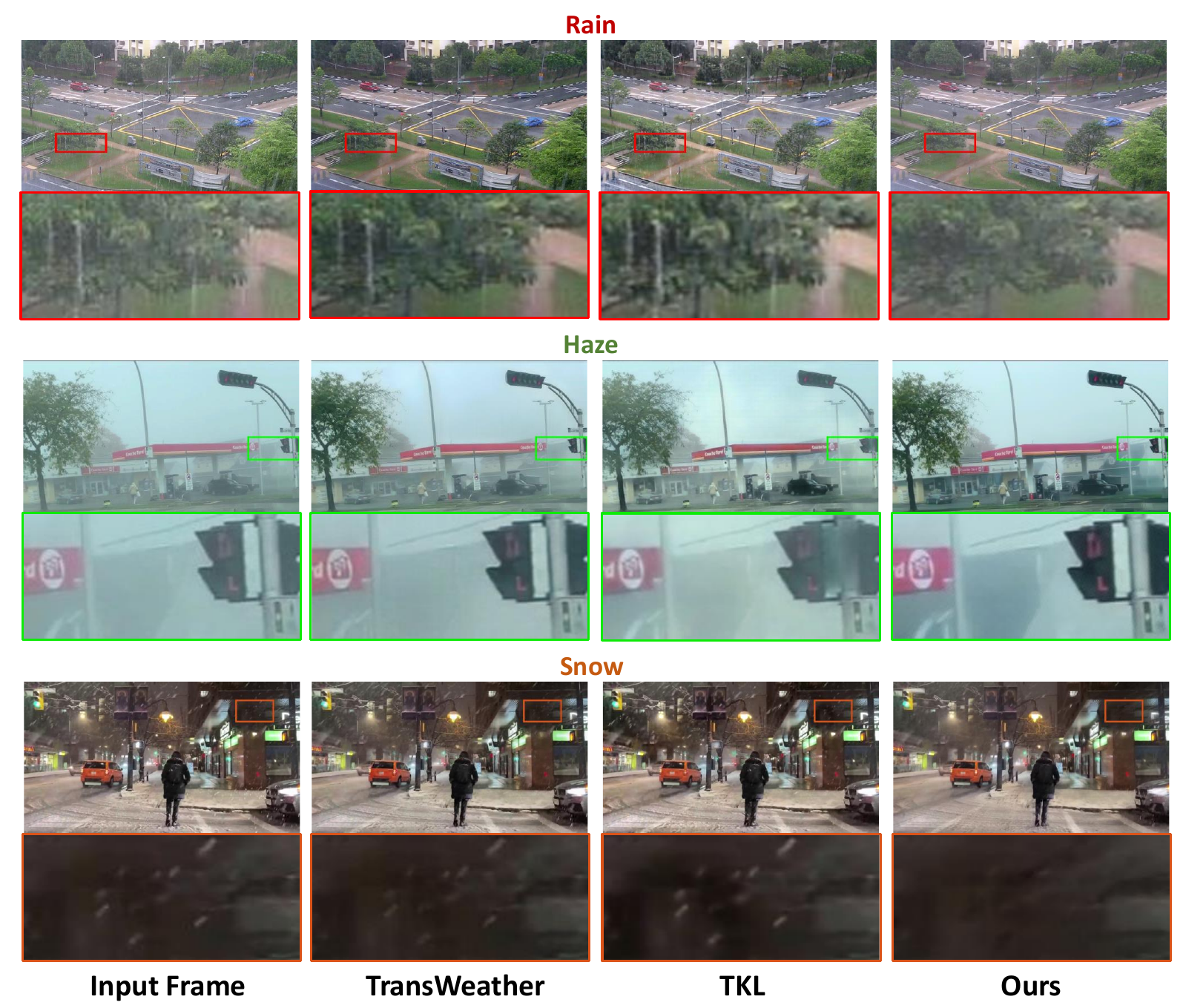}
\caption{\textbf{Visual comparison of different multiple adverse weather removal methods on three real-world video sequences degraded by rain, haze, snow, respectively.} The color boxes display zoom-in views highlighting detailed comparisons of weather removal. Apparently, our network can more effectively remove rain streaks, haze, and snowflakes of input video frames than state-of-the-art methods.}
\label{fig:realworld}
\end{figure}

\begin{table}[h]
  \centering
  \caption{Quantitative comparison of computational complexity between the selected models and ViWS-Net. The best values are denoted in bold.}
    % \vspace{-0.4cm}
    \resizebox{\linewidth}{!}{%
    \begin{tabular}{c|c|c|c}
    \toprule
    \textbf{Methods} & \textbf{Parameters} (M) & \textbf{FLOPs} (G) & \textbf{Inference time} (s) \\ 
    \hline
    \hline
    TransWeather~\cite{valanarasu2022transweather} & 24.01 & \textbf{37.68} & 0.49  \\
    %\hline
    TKL~\cite{chen2022learning}  & 28.71 & 94.05 & 0.51  \\
    EDVR~\cite{wang2019edvr}  & \textbf{20.70} & 335.27 & 0.63  \\
    \hline
    \textbf{ViWS-Net}(Ours)  & 57.82 & 68.72 & \textbf{0.46}  \\
    \bottomrule
    \end{tabular}
    }
  \label{tab:complexity}
\end{table}

\begin{table*}[!tbp]
\centering
  \caption{\textbf{Ablation study of each critical module in the proposed framework on three weather types.} The top values are marked in bold font. ``WS. Adv.'' denote weather-suppression adversarial learning.}
    \resizebox{1.0\textwidth}{!}{%
\begin{tabular}{c|ccc|cc|cc|cc|cc} 
\toprule
\multirow{2}{*}{\textbf{Combination}} & \multicolumn{3}{c|}{\textbf{Module}}                                      & \multicolumn{8}{c}{\textbf{Datasets}}                                                                            \\ 
\cline{2-12}
                                      & \textbf{WeatherMessenger} & \textbf{VideoDecoder} & \textbf{WS. Adv.} & \multicolumn{2}{c|}{Rain} & \multicolumn{2}{c|}{Haze} & \multicolumn{2}{c|}{Snow} & \multicolumn{2}{c}{Average}  \\ 
\hline
\hline
M1                              & \textbf{-}                         & \textbf{-}                     & \textbf{-}                     & 26.92 & 0.9273            & 22.77 & 0.9052            & 29.94 & 0.9462            & 26.54 & 0.9262               \\
M2                                    & \checkmark                         & \textbf{-}                     & \textbf{-}                     & 30.03 & 0.9327            & 23.92 & 0.9149            & 30.54 & 0.9520            & 28.16 & 0.9332               \\
M3                                    & \textbf{-}                         & \checkmark                     & \textbf{-}                     &    29.33   &   0.9365                &    22.84   &       0.9085            &   30.89    &     0.9554              &  27.69     &     0.9335                 \\
M4                                    & \textbf{-}                         & \textbf{-}                     &\checkmark                     & 29.70 & 0.9316            & 23.87 & 0.9152            & 30.82 & 0.9521            & 28.13 & 0.9330               \\
M5                                    & \checkmark                         &\checkmark                     & \textbf{-}                     & 31.00 & 0.9419            & 24.13 & 0.9164            & 30.93 & 0.9552            & 28.69 & 0.9378               \\
\textbf{Ours}                         &\checkmark                         & \checkmark                     & \checkmark                     &    \textbf{31.52}   &    \textbf{0.9433}               &    \textbf{24.51}   &    \textbf{0.9187}               &    \textbf{31.49}   &    \textbf{0.9562}               &   \textbf{29.17}    &    \textbf{0.9394}                  \\
\bottomrule

\end{tabular}
    }
\label{tab:ablation}
\end{table*}

\begin{table}
\centering
\caption{\textbf{Ablation study of the proposed messenger-driven video transformer decoder.} The top values are denoted in bold.}
\begin{tabular}{cc|cc} 
\toprule
\multicolumn{2}{c|}{\textbf{Module}}         & \multicolumn{2}{c}{\textbf{Average}}    \\ 
\hline
TemporalFusion & RefineNet & PSNR & SSIM  \\ 
\hline
\hline
   \textbf{-}                     &      \textbf{-}              &         28.37 	       &      0.9305          \\
\checkmark                       &         \textbf{-}           &       28.80 	         &     0.9357           \\
\checkmark                       & \checkmark                  &    \textbf{29.17}           &     \textbf{0.9394}          \\
\bottomrule
\end{tabular}
\label{tab:decoder}
\end{table}

\subsection{Quantitative Evaluation}
\noindent \textbf{Comparison methods. }
As shown in Table~\ref{tab:main}, we compared our proposed method against five kinds of state-of-the-art methods on our mixed dataset. For \textit{derain}, we compared our method with one single-image approach PReNet~\cite{ren2019progressive} and three video approaches SLDNet~\cite{yang2020self}, S2VD~\cite{yue2021semi}, RDD-Net~\cite{wang2022rethinking}.
For \textit{dehaze}, we compared with two single-image approaches GDN~\cite{liu2019griddehazenet}, MSBDN~\cite{dong2020multi} and two video approaches VDHNet~\cite{ren2018deep}, PM-Net~\cite{liu2022phase}. For \textit{desnow}, we compared with four single-image methods including DesnowNet~\cite{liu2018desnownet}, DDMSNET~\cite{zhang2021deep}, HDCW-Net~\cite{chen2021all}, SMGARN~\cite{cheng2022snow}. For \textit{restoration}, we compared ours with one single-image method MPRNet~\cite{zamir2021multi} and three video methods EDVR~\cite{wang2019edvr}, RVRT~\cite{liang2022recurrent}, RTA~\cite{zhou2022revisiting}. For \textit{multi-adverse-weather removal}, we compared ours with the latest four single-image methods All-in-one~\cite{li2020all}, UVRNet~\cite{kulkarni2022unified}, TransWeather~\cite{valanarasu2022transweather}, TKL~\cite{chen2022learning}.

\noindent \textbf{Analysis on multi-adverse-weather removal. }
For quantitative evaluation of the restored results, we apply the peak signal-to-noise ratio (PSNR) and the structural similarity (SSIM) as the metrics.
For the single-weather removal models (derain, dehaze, desnow), two types of results are reported: \textbf{(i)} the model trained on their original weather (i.e., single weather training set) and \textbf{(ii)} the model trained on data of all weather types (i.e., the mixed training set). For restoration and multi-adverse-weather removal models, only the results of the model trained on the mixed training set are reported.
For a fair comparison, we retrain each compared model implemented by the official codes based on our training dataset and report the best result.
One can see that, our method achieves the best average performance when trained on multi-weather types by a considerable margin of 2.16, 0.0216 in PSNR, SSIM, respectively, than the second-best method EDVR~\cite{wang2019edvr}. Although our method may not be the best compared to single-weather removal methods when trained on single-weather data, these methods usually go to failure when coming to multiple adverse weather conditions. For example, while the derain method RDD-Net~\cite{wang2022rethinking} fails to remove the haze degradation, the dehaze method PM-Net~\cite{liu2022phase} and desnow method DDMSNET~\cite{zhang2021deep} have poor performance on snow and haze removal, respectively.
Also, it can be observed that DDMSNET~\cite{zhang2021deep} and SMGARN~\cite{cheng2022snow} still achieve promising results for snow removal when trained on multi-weather types by incorporating snow-specialized modules. However, these methods struggle to address other degradations like haze, leading to lower average performance in multi-weather restoration. 
% In contrast, we emphasize consistent performance across all weather types.
%
% In contrast, our method can achieve consistent performance compared to existing methods when we address all weather types by solely adopting a set of pre-trained weights and a unified architecture.
In contrast to existing methods, our approach can achieve consistent performance across all weather types by relying solely on a unified architecture and a set of pre-trained weights.

\noindent \textbf{Analysis on computational complexity. }
We evaluate computational complexity (the number of training parameters, FLOPs, inference time) by feeding a 5-frame video clip with a resolution of 224$\times$224 into our model and the representative models. Our ViWS-Net maintains comparable computational complexity to other methods while achieving the best results on multi-adverse-weather removal.

\subsection{Qualitative Evaluation}
\noindent \textbf{Results on our datasets.}
To better illustrate the effectiveness of our ViWS-Net, Figure~\ref{fig:vis_comparison} shows the visual comparison under our rain, haze, and snow scenarios between our method and 5 state-of-the-art methods that are, respectively, the one with the best average performance for each group of methods. Obviously, one can notice that our method can achieve promising results in visual quality in each weather type. For rain and snow scenarios, the results recovered by our method contain less rain streaks and snow particles compared with other methods. For the hazy scenario, our method can remove more residual haze and much better preserve clean background.

\vspace{3pt}\noindent \textbf{Results on real-world degraded videos.}
To evaluate the universality of our video multiple adverse weather removal network, we collect three real-world degraded videos, \ie, one rainy video from NTURain~\footnote{https://github.com/hotndy/SPAC-SupplementaryMaterials/}, one hazy video and one snowy video from Youtube website, and further compare our network against state-of-the-art multi-adverse-weather removal methods. Figure~\ref{fig:realworld} shows the visual results produced by our network and two selected methods on real-world video frames. Apparently observed from the detailed comparison, our method outperforms other methods in all weather types by effectively removing adverse weather and maintaining background details. 
%Although TransWeather and TKL may remove parts of haze in rainy and snowy videos, they failed to eliminate the main adverse weather, \ie, rain and snow respectively.

\subsection{Ablation Study}
\noindent \textbf{Effectiveness of each module in ViWS-Net.}
We evaluate the effectiveness of each proposed module including temporally-active weather messenger, video transformer decoder, and weather-suppression adversarial learning (WS. Adv.) as shown in Table~\ref{tab:ablation}. We report the result tested on the weather-specific testing set and trained on the mixed training set. The baseline M1, which consists of a Shunted Transformer encoder and a convolution projection decoder, achieves the average performance on three adverse weather datasets of 26.54, 0.9262 in PSNR, SSIM, respectively. M2 introduces temporally-active weather messenger tokens in the transformer encoder based on M1 and advances the average performance by 1.62, 0.0070 of PSNR, SSIM, respectively, demonstrating the effectiveness of our proposed Long Short-term Temporal Modeling strategy. M3 presents the messenger-driven video transformer decoder (weather type queries are randomly initialized), while M4 brings in the weather-suppression adversarial learning based on M1. Both M3 and M4 boost the average performance by a significant margin. M5 is developed from M2 and M3, where the weather type queries are initialized by the well-learned weather messenger tokens, leading to a better average performance of 28.69, 0.9378 in PSNR, SSIM. Our full model further applies the weather-suppression adversarial learning strategy and gains a critical increase of 0.48, 0.0016 in PSNR, SSIM, respectively, compared with M5.

\vspace{3pt}\noindent \textbf{Effectiveness of video transformer decoder.}
We further validate the effectiveness of Temporal Fusion module and RefineNet module in our elaborated video transformer decoder as shown in Table~\ref{tab:decoder}. Our reported results were obtained by testing our approach on a mixed testing set and training it on a mixed training set.
It is worth noting that both of them benefit the average performance.

%%%% Conclusion
\section{Conclusion}

This paper presents ViWS-Net, an innovative method for simultaneously addressing multiple adverse weather conditions in video frames using a unified architecture and a single set of pre-trained weights. Our approach incorporates Weather-Suppression Adversarial Learning to mitigate the adverse effects of different weather conditions, and Weather Messenger to leverage rich temporal information for consistent recovery. We evaluate our proposed method on benchmark datasets and real-world videos, and our experimental results demonstrate that ViWS-Net achieves superior performance compared to state-of-the-art methods. Ablation studies are also conducted to validate the effectiveness of each proposed module.

\section*{Acknowledgments}

This work was supported by the Guangzhou Municipal Science and Technology Project (Grant No. 2023A03J0671),
National Natural Science Foundation of China (Grant No. 61902275), and
Hong Kong Metropolitan University Research Grant (No. RD/2021/09).

%\clearpage
{\small
\bibliographystyle{ieee_fullname}
\bibliography{egbib}
}
\end{document}